%% file: main.tex
\documentclass[twoside]{article}
\pdfoutput=1

\usepackage[accepted]{aistats2023}
\usepackage{hyperref}
\hypersetup{hidelinks=true}
\usepackage{apacite}
\usepackage{changes}
\usepackage{amsmath,amssymb}
\usepackage{dsfont}
\usepackage{mathtools}
\usepackage{algorithm}
\usepackage{algpseudocode}
\usepackage{booktabs}
\usepackage{subcaption}
\input{tables}

\input{figures}

\input{algorithms}
\input{math_commands}

\begin{document}


\runningauthor{Yang Yang, Gennaro Gala, Robert Peharz}

\twocolumn[

\aistatstitle{Bayesian Structure Scores for Probabilistic Circuits}

\aistatsauthor{Yang Yang\textsuperscript{*} \And Gennaro Gala\textsuperscript{*} \And  Robert Peharz}

\aistatsaddress{Eindhoven University of Technology \\ KU Leuven \And Eindhoven University of Technology \And Graz University of Technology \\ Eindhoven University of Technology} ]

\begin{abstract}
Probabilistic circuits (PCs) are a prominent representation of probability distributions with tractable inference.  
While parameter learning in PCs is rigorously studied, structure learning is often more based on heuristics than on principled objectives.
In this paper, we develop Bayesian structure scores for deterministic PCs, i.e., the structure likelihood with parameters marginalized out, 
which are well known as rigorous objectives for structure learning in probabilistic graphical models.
When used within a greedy cutset algorithm, our scores effectively protect against overfitting and yield a fast and almost hyper-parameter-free structure learner, distinguishing it from previous approaches.
In experiments, we achieve good trade-offs between training time and model fit in terms of log-likelihood.
Moreover, the principled nature of Bayesian scores unlocks PCs for accommodating frameworks such as structural expectation-maximization.
\end{abstract}

\section{INTRODUCTION}
Probabilistic circuits (PCs) \shortcite{vergari2020probabilistic} have emerged as a powerful framework for describing tractable probabilistic models, such as \emph{Chow-Liu trees} \shortcite{chow1968approximating}, \emph{arithmetic circuits} \shortcite{darwiche2003differential}, \emph{sum-product networks} \shortcite{poon2011sum}, \emph{probabilistic sentential decision diagrams} \shortcite{kisa2014probabilistic} and \emph{cutset networks} (CNets) \shortcite{rahman2014cutset}. 
PCs can be categorized into several sub-families specified by \emph{structural properties} or \emph{constraints}, which go hand in hand with certain tractable inference routines \shortcite{vergari2020probabilistic}, such as \emph{marginalization}, \emph{conditioning}, \emph{most-probable explanation}, \emph{expectations}, etc.

PCs can either be compiled from other models \shortcite{darwiche2003differential} or learned directly from data \shortcite{Lowd2008,poon2011sum}.
When learning PCs from data, we might---similar as in classical probabilistic graphical models (PGMs) \shortcite{koller2009probabilistic}---distinguish between \emph{parameter learning} and \emph{structure learning}.
Parameter learning has been rigorously studied in PCs and mirrors state-of-the-art in PGMs.
Specifically, techniques for parameter learning in PCs are \emph{(closed-form) maximum-likelihood estimation} (for the sub-class of deterministic PCs) \shortcite{kisa2014probabilistic,peharz2014learning}, \emph{expectation-maximization} \shortcite{peharz2016latent}, \emph{concave-convex procedures} \shortcite{zhao2016unified}, \emph{Bayesian approaches} \shortcite{zhao2016collapsed,rashwan2016online,vergari2019automatic,trapp2019bayesian}, \emph{discriminative methods} \shortcite{rashwan2018discriminative, peharz2020random}, \emph{continuous embeddings} \shortcite{shao2022conditional,correia2022continuous}, and \emph{latent variable distillation} \shortcite{liu2022scaling}.

\emph{Structure learning} in PCs, however, is generally less principled than in PGMs.
In the latter, the two main styles of structure learning are \emph{constraint-based}, i.e., detecting conditional independencies in the data and translating them into a corresponding graph structure \shortcite{koller2009probabilistic}; and \emph{score-based}, i.e., phrasing structure learning as a discrete optimization problem, aiming to find a graph that maximizes some structure score.
Score-based approaches have turned out to be more practical, as constraint-based approaches tend to be sensitive to statistical noise in conditional independence tests \shortcite[page~790]{koller2009probabilistic}.
In particular, \emph{Bayesian structure scores} \shortcite{buntine1991theory,geiger1994learning,heckerman1995learning} are some of the most prominent scores, due to their principled nature and asymptotic consistency guarantees \shortcite{koller2009probabilistic}.
Bayesian scores are the \mbox{(log-)likelihood} of the candidate structures, with parameters marginalized out, and have been the target of both greedy \shortcite{cooper1992bayesian} and exact \shortcite{de2009structure} structure optimizers.

In PCs, however, structure learners are often based on heuristics.
The most prominent structure learning scheme is probably \emph{LearnSPN} \shortcite{gens2013learning}, which, in a nutshell, uses top-down multi-clustering to lay out the PC structure.
The same approach is taken in \emph{ID-SPN} \shortcite{rooshenas2014learning}, which augments LearnSPN by using expressive distribution models as sub-modules.
Other structure learners are based on singular-value decomposition \shortcite{adel2015learning}, bottom-up information bottleneck learning \shortcite{peharz2013greedy} or are motivated by decision tree learning \shortcite{rahman2014cutset}.
None of these approaches declare an \emph{explicit objective} for structure learning, although at test time they are usually evaluated in terms of log-likelihood.
Moreover, they often rely on a large number of hyper-parameters, which are tedious to set or need to be tuned on a separate validation set.

There are also score-based approaches to learn PC structures, e.g.~the ones by \shortciteA{Lowd2008,peharz2014learning}, who use a weighted sum of training log-likelihood and circuit size as a structure score.
The circuit size directly corresponds to the worst-case inference cost in PCs, thus these scores have the interesting interpretation of trading off ``model fit'' vs.~``inference complexity.'' 
However, a deeper theoretical justification of these scores has yet to be shown.
Similarly, the \emph{Strudel} method \shortcite{dang2020strudel} also performs a structure search based on ``\emph{heuristic scores calculated from data}.''
Thus, these score-based methods are---similarly as the previously mentioned structure learners---of heuristic nature, usually equipped with various (often unintuitive) hyper-parameters for regularizing the model, and require a separate validation set.

In this paper, we propose a principled avenue to PC structure learning and put a particular emphasis on elegance and simplicity.
Specifically, we derive \emph{Bayesian structure scores} for \emph{deterministic} PCs,\footnote{Extensions to general, non-deterministic PCs are left to future work, but possible directions are given in the Discussion.} by equipping parameters with suitable priors and marginalizing them from the Bayesian model.
Thus, similarly as in PGMs, a Bayesian score is the marginal likelihood of the PC structure under a particular choice of parameter prior.
Intuitively, since the Bayesian score averages over \emph{all} possible choices of parameters, it effectively protects against overfitting, as overly complex structures will have a significant portion of ``bad'' parameters that are not well-supported by the data.

In the special case of discrete data and Dirichlet priors, one yields the well-known \emph{Bayes-Dirichlet} (BD) score \shortcite{buntine1991theory,heckerman1995learning}.
We employ the BD score within \emph{greedy cutset learning} \shortcite{rahman2014cutset}, a simple and fast structure learning scheme for PCs.
Our method has a single hyper-parameter governing the Dirichlet prior, the \emph{equivalent sample size} (ESS), which we, however, keep fixed to $0.1$ throughout our experiments, rendering our method effectively \emph{hyper-parameter-free}.

Additionally, we consider the \emph{Bayesian information criterion} (BIC), which is frequently used as an alternative to the BD score, as they are asymptotically equivalent \shortcite{schwarz1978estimating,koller2009probabilistic}.
The BIC score has indeed been considered for PCs \shortcite{dimauro2015learning}, but in a form which undercounts the actual number of parameters.
Our corrected version of the BIC score improves the results by \shortciteA{dimauro2015learning} and is demonstrated to be a viable and efficient alternative to our full BD score.

Both BD and BIC are de facto \emph{hyper-parameter-free} and designed to reflect generalization.
Therefore, they can use the full training data, without the need to dedicate part of it to a validation set. 
Both scores deliver models competitive with state-of-the-art when learned on 20 common benchmark density estimation data sets, even though we are using a simplistic structure learner.
We compare our approach with various other PC structure learners, both simple ones---which we run, like our methods, with their default hyper-parameters on the entire training data---and sophisticated ones---which use extensive tuning but consume orders of magnitude more computational resources.
Our structure learners are often located on the Pareto front determined by training time and test log-likelihood, i.e.~they are among the best fast structure learners.
This, together with the fact that we use well-principled structure objectives, make our algorithms highly practical.

Moreover, since a Bayesian score is just the marginal structure likelihood, we can naturally embed our structure learner within a \emph{structural expectation maximization (EM)} algorithm \shortcite{DBLP:conf/uai/Friedman98} for learning mixtures of PCs.
Structural EM, which to the best of our knowledge has not been applied to PCs before, consistently improves over single PCs learned with BD, and delivers models close to or surpassing state-of-the-art.

The source code to replicate the experiments is available at \url{https://github.com/yangyang-pro/bayesian-scores-pc}.

\section{PROBABILISTIC CIRCUITS}
\label{sec:pcs}

Given a set of random variables $\X$, a probabilistic circuit (PC) over $\X$
is a computational (directed acyclic) graph $\graph=(V, E)$ containing three types of computational nodes, namely \emph{distribution nodes}, \emph{weighted sums} and \emph{products}.
As we elaborate below, each node $\node \in V$ computes a (possibly unnormalized) probability distribution over some subset of $\X$, denoted as the \emph{scope} of $\node$.
Hence, we associate with each PC a \emph{scope function} $\scope \colon V \to 2^{\X}$, assigning to each node $\node\in V $ its scope $\scope(\node) \subseteq \X$.
Naturally, the scope of any internal node $\node$ satisfies $\scope(\node)=\bigcup_{\node' \in \inputs(\node)}\scope(\node')$, where $\inputs(\node)$ is the set of all input nodes to $\node$, thus the scope function $\sigma$ is completely specified by the scopes of the leaves (also called input nodes) of $\graph$.

Any leaf $\leaf \in V$ of the PC is a probability distribution over its scope $\scope(\leaf)$, where one typically uses some ``simple'' parametric distribution like Gaussian, Poisson, categorical, etc., whose parameters are denoted as $\theta_\leaf$.
The internal nodes of the PC are either sums or products.
A sum node $\sumnode$, equipped with weights $\bm{w}_{\sumnode}=\{w_{\sumnode\node}\}_{\node \in \inputs(\sumnode)}$, computes a convex combination (mixture) of its inputs, i.e., \mbox{$\sumnode = \sum_{\node \in \inputs(\sumnode)} w_{\sumnode\node} \, \node$}, where $\sum_{\node \in \inputs(\sumnode)} w_{\sumnode\node}=1$ and $w_{\sumnode\node} \geq 0$;
a product node $\prodnode$ computes the product of its inputs, i.e., $\prodnode = \prod_{\node \in \inputs(\prodnode)}\node$.
The parameters $\{\Theta, \bm{w}\}$ of the PC are all parameters of the leaf distributions \mbox{$\Theta = \{\theta_\leaf\}_{\leaf \in V}$} and all sum weights $\bm{w} = \{\bm{w}_\sumnode\}_{\sumnode \in V}$, where the expression $\leaf \in V$ (respectively $\sumnode \in V$) means that we range over all leaves (respectively sum nodes) in $V$.

Since all leaves are probability distributions and all sum-weights are non-negative, it follows that each node in $V$ computes some (possibly unnormalized) distribution over its scope.
We assume that the PC has a single output node, having full scope $\X$, which represents the model distribution.
It can be computed for any input sample $\x$ using a feed-forward pass in the PC.

Besides evaluating the model density, PCs also allow a wide range of tractable probabilistic inference routines, if they satisfy certain structural properties (or constraints), denoted as \emph{decomposability}, \emph{smoothness}, \emph{determinism}, and \emph{structured decomposability}.
Specifically:
\begin{itemize}
    \item A PC is \emph{decomposable}, if for each product node $\prodnode$ it holds that $\scope(\node) \cap \scope(\node') = \emptyset$, for all $\node \neq \node' \in \inputs(\prodnode)$.
    \item A PC is \emph{smooth}, if for each sum node $\sumnode$ it holds that $\scope(\node) = \scope(\node')$, for all $\node, \node' \in \inputs(\sumnode)$.
    \item A PC is \emph{deterministic}, if for each sum node $\sumnode$ and each sample $\x$, at most one of the inputs of $\sumnode$ is non-zero.
    \item Furthermore, \emph{structured decomposability} denotes a stricter form of decomposability, where the decompositions of all products adhere to a common pattern described by a so-called \emph{vtree} \shortcite{pipatsrisawat2008new, kisa2014probabilistic, ahmed2022semantic}.
\end{itemize}

Not all properties are needed for each and every inference scenario, and studying the correspondence between structural properties and tractable inference routines is one of the intriguing facets of PCs \shortcite{darwiche2002knowledge,vergari2020probabilistic}.
Typically one assumes at least the properties decomposability and smoothness, which ensure that each probability distribution computed by any node $\node \in V$ is already normalized.
More importantly, they enable tractable \emph{marginals} and \emph{conditionals}---the two core routines of probabilistic reasoning \shortcite{ghahramani2015probabilistic}---in time linear in the circuit size \shortcite{Peharz2015}.
In the remainder of the paper, we assume that any PC is decomposable and smooth.

Adding determinism yields tractable inference of \emph{most-probable explanations} (MPE), i.e., finding maximizers of the PC distribution, and exact computation of \emph{maximum-likelihood parameters} \shortcite{kisa2014probabilistic,peharz2014learning}.
In this paper, we will see that determinism also allows to compute Bayesian structure scores exactly and efficiently.

Structured decomposability allows taking expectations of one circuit with respect to another \shortcite{khosravi2019expect}, computing divergences between PCs \shortcite{vergari2021compositional}, etc.

PCs have evolved into a ``lingua franca'' of tractable models, since many of them can be cast into the PC framework.
PCs might be compiled from other models \shortcite{darwiche2003differential} or learned directly from data, where we can distinguish between parameter learning and structure learning.
As mentioned in the introduction, structure learning in PCs is usually rather ad-hoc and heuristic.
In this paper, we develop \emph{Bayesian scores} as well-principled objectives for PC structure learning.

\section{BAYESIAN STRUCTURE SCORES FOR DETERMINISTIC PCS}
\label{sec:BayesianScores}

Assume a training set $\data = \{\x^{(n)}\}_{n=1}^{N}$ of $N$ samples drawn i.i.d.~from an unknown probability distribution, and a structured, parametric density model $p(\x \cbar \Theta, \graph)$, that is, a distribution over $\X$ conditional on continuous parameters $\Theta$ and a structure parameter $\graph$.
The basic idea of the \emph{Bayesian structure score} is to (i) equip $\Theta$ with some prior distribution $p(\Theta \cbar \graph)$, and to (ii) derive the marginal likelihood of $\graph$ by marginalizing the parameters from the Bayesian model:
\begin{equation}    \label{eq:Bayes_score}
\mathcal{B}(\graph)
\coloneqq 
p(\data \cbar \graph) 
= \int p(\Theta \cbar \graph) \, \prod_{\x \in \data} p(\x \cbar \Theta, \graph) \, \mathrm{d}\Theta.
\end{equation}
The Bayesian score might be used as a direct target of optimization, i.e.~a maximum likelihood approach with respect to $\graph$.
Alternatively, it might also be wrapped into a larger probabilistic framework, e.g.~a Bayesian approach with respect to structure \shortcite{friedman2003being} or structural expectation maximization \shortcite{DBLP:conf/uai/Friedman98}.

While eq.~(\ref{eq:Bayes_score}) is a hard computational problem in general, there are interesting special cases where it can be computed exactly and efficiently.
In particular, in the context of Bayesian networks over discrete data, the well-known \emph{Bayes-Dirichlet} (BD) score has been studied by \shortciteA{buntine1991theory, cooper1992bayesian, heckerman1995learning}.
By using Dirichlet priors on the parameters, together with particular assumptions such as independence among variable families, parameter modularity among different structures, etc., the BD score can be computed analytically in Bayesian networks.
Similarly, for Bayesian networks with Gaussian parameters and normal-Wishart priors, the \emph{Bayes-Gauss} (BG) score has been developed \shortcite{geiger1994learning}.

Let now $\graph = (V,E)$ be a candidate PC structure with parameters 
$\{\Theta, \bm{w}\}$, where \mbox{$\Theta = \{\theta_\leaf\}_{\leaf \in V}$} and $\bm{w} = \{\bm{w}_\sumnode\}_{\sumnode \in V}$.
The parameters of each leaf $\leaf$ are equipped with a prior $p(\theta_\leaf)$, where we assume that eq.~(\ref{eq:Bayes_score}), when restricted to $\leaf$, can be computed exactly and efficiently.
That is, we can compute 
\begin{equation}
\label{eq:leaf_Bayes}
    \mathcal{B}_\leaf = \int 
    p(\theta_\leaf) \,
    \prod_{\x \in \data} p(\x \cbar \theta_\leaf) \, \mathrm{d}\theta_\leaf.
\end{equation}
In order to keep the notation uncluttered, we write $p(\x \cbar \theta_\leaf)$, which should be understood as an evaluation of the leaf on the sub-vector of $\x$ whose entries correspond to $\scope(\leaf)$.
Eq.~(\ref{eq:leaf_Bayes}) can be computed in closed form for many exponential families with conjugate priors, e.g.~the Gaussian-Wishart and Binomial-Dirichlet families.

Further, let $\sumnode$ be an arbitrary sum node with weights 
\mbox{$\bm{w}_{\sumnode}=\{w_{\sumnode\node}\}_{\node \in \inputs(\sumnode)}$}, 
which we equip with a Dirichlet prior
\begin{equation}
p(\bm{w}_\sumnode) = {1 \over B(\bm{\alpha}_{\sumnode})} \prod_{\node \in \inputs(\sumnode)} w_{\sumnode \node}^{\alpha_{\sumnode\node}-1},
\end{equation}
parameterized by $\bm{\alpha}_{\sumnode} = \{\alpha_{\sumnode\node}\}_{\node \in \inputs(\sumnode)}$, where $\alpha_{\sumnode\node} > 0$, and $B(\bm{\alpha}_{\sumnode})$ is the Beta function, i.e. the normalization constant of the Dirichlet distribution.  
Assuming a-priori parameter independence, we get the prior
\begin{align}   
    p(\Theta, \bm{w} \cbar \graph) 
    & = \prod_{\sumnode \in V} p(\bm{w}_\sumnode) \prod_{\leaf \in V} \, p(\theta_\leaf)  \nonumber
    \\
    & = \prod_{\sumnode \in V} {1 \over B(\bm{\alpha}_{\sumnode})}  \prod_{\node \in \inputs(\sumnode)}  w_{\sumnode \node}^{\alpha_{\sumnode\node}-1}  \prod_{\leaf \in V} \, p(\theta_\leaf). 
    \label{eq:factorized_prior}
\end{align}
The structure $\graph$, leaf parameters $\Theta$, and sum-weights $\bm{w}$ completely parametrize the PC distribution $p(\x \cbar \Theta, \bm{w}, \graph)$,\footnote{For simplicity, we leave the scope function implicit, i.e.~we assume that each leaf ``knows'' its scope.} 
such that the Bayesian structure score for PCs is given as
\begin{equation}      \label{Bayes_score_PC}
\mathcal{B}_{PC}(\graph)
= \int \int 
p(\Theta, \bm{w} \cbar \graph) \,
\prod_{\x \in \gD} p(\x \cbar \Theta, \bm{w}, \graph) \, \mathrm{d}\Theta \, \mathrm{d}{\bm{w}}.
\end{equation}
While computing $\mathcal{B}_{PC}(\graph)$ is intractable for \emph{non-deterministic} PCs, due to the high-dimensional integrals which do not simplify further, we can derive an analytic solution for \emph{deterministic} PCs.

Recall that in deterministic PCs, for each possible sample $\x$ and each sum node $\sumnode$ at most one input to $\sumnode$ is non-zero.
In general, whether a PC is deterministic or not might depend on both the structure $\graph$ and the parameters $\{\Theta,\bm{w}\}$ in a convoluted manner.
In this paper, however, we require that for given $\graph$ determinism holds for each choice of parameters, a property which we call \emph{structural determinism}.
Moreover, we require that the indices of the non-zero sum inputs remain unchanged when varying parameters, i.e., for each sample $\x$ and sum node $\sumnode$ it holds that if $\sumnode$ has a non-zero input under two distinct parameters $\{\Theta',\bm{w}'\}$ and $\{\Theta'',\bm{w}''\}$, then the same input must be non-zero for both parameter choices.
We call this property \emph{parameter-consistent determinism}, which is often assumed by default in literature \shortcite{choi2020probabilistic}.

A simple way to enforce structural and parameter-consistent determinism is to associate with each sum $\sumnode$ a corresponding random variable $X_\sumnode \in \X$, and let $\sumnode$ be of the form
\begin{equation}
\sumnode = \sum_k p_k(X_{\sumnode}) \times p_k(\scope(\sumnode) \setminus X_{\sumnode}),
\label{eq:decision}
\end{equation}
where $p_k(X_{\sumnode})$ are distributions over $X_\sumnode$ (leaves of the PC) whose supports\footnote{The \emph{support} is the set of inputs where the density is non-zero.} are disjoint and independent of any parameters, and $p_k(\scope(\sumnode) \setminus X_{\sumnode})$ are arbitrary PC nodes with scope $\scope(\sumnode) \setminus X_{\sumnode}$.
This form of determinism is very common and known under various names such as \emph{decision} \shortcite{darwiche2002knowledge} or \emph{regular selectivity} \shortcite{peharz2014learning}.

The key to computing the Bayes score in deterministic PCs is the \emph{tree decomposition} introduced by \shortciteA{zhao2016collapsed}.
Specifically, they introduced the notion of \emph{induced tree}, a sub-graph of $\graph$ obtained by (i) removing all inputs except one from each sum node in $\graph$, and (ii) removing all nodes which are rendered unreachable from the root.
Since we can decide for each sum node independently which input to remove, the number of induced trees is exponential in the number of sum nodes.
They show that the PC distribution can be written as a ``large flat mixture,'' the so-called tree decomposition (cf.~also \shortciteA{trapp2019bayesian})
\begin{equation}
\label{eq:tree_decomposition}
p(\x \cbar \Theta, \bm{w}, \graph) = 
\sum_{\tau \in \graph} 
\prod_{\sumnode\node \in E}  w_{\sumnode \node}^{\mathds{1}[\sumnode \node \in \tau]}
\prod_{\leaf \in V} p(\x \cbar \theta_\leaf)^{\mathds{1}[\leaf \in \tau]},
\end{equation}
where the sum runs over all induced trees contained in $\graph$, the first product runs over all sum-input edges, and the second product over all leaves.
Here, $\mathds{1}[\sumnode \node \in \tau]$ is the indicator that the edge between $\sumnode$ and $\node$ is contained in $\tau$ and $\mathds{1}[\leaf \in \tau]$ indicates whether $\leaf$ is in $\tau$.
Thus, each sum term in (\ref{eq:tree_decomposition}) is the product of all sum-weights and leaves appearing in $\tau$.

It turns out that, for some parameter set of non-zero measure under prior (\ref{eq:factorized_prior}), there is for each $\x \in \data$ \emph{exactly one} non-zero term in the exponential sum in eq.~(\ref{eq:tree_decomposition}), corresponding to a particular induced tree $\tau_{\x}$.
First, we can assume without loss of generality that there exists \emph{at least} one non-zero term---otherwise (\ref{Bayes_score_PC}) would evaluate to zero, which would be the correct score for the considered candidate structure.
Further, in decomposable and deterministic PCs,
there can be \emph{at most} one non-zero term, corresponding to a particular induced tree $\tau_{\x}$.
This induced tree can be found by tracing the PC backwards from root to leaves, where at sum nodes the single non-zero input is followed and at product nodes all inputs are followed.
If at any sum node a zero-input was selected, it can be shown that the corresponding term in eq.~(\ref{eq:tree_decomposition}) evaluates to zero \shortcite{peharz2014learning}.
Consequently, eq.~(\ref{eq:tree_decomposition}) reduces to \begin{equation}
\label{eq:tree_decomposition_deterministic}
p(\x \cbar \Theta, \bm{w}, \graph) = 
\prod_{\sumnode\node \in E} w_{\sumnode \node}^{\mathds{1}[\sumnode \node \in \tau_\x]}
\prod_{\leaf \in V} p(\x \cbar \theta_\leaf)^{\mathds{1}[\leaf \in \tau_\x]}.
\end{equation}
Substituting (\ref{eq:factorized_prior}) and (\ref{eq:tree_decomposition_deterministic}) into (\ref{Bayes_score_PC}) and re-arranging products yields
\begin{align}
\mathcal{B}_{PC}(\graph) 
\!
= 
\!\!
\int \!\!\! \int  \!
& \Bigg( 
\prod_{\sumnode \in V} 
{1 \over B(\bm{\alpha}_{\sumnode})} \! \prod_{\node \in \inputs(\sumnode)} \!\!\! w_{\sumnode \node}^{\alpha_{\sumnode\node}-1} 
\prod_{\x \in \data} \!
w_{\sumnode\node}^{\mathds{1}[\sumnode \node \in \tau_\x]} 
\Bigg) 
\nonumber \\
& \Bigg( \prod_{\leaf \in V} p(\theta_\leaf) \prod_{\x \in \data} p(\x \cbar \theta_\leaf)^{\mathds{1}[\leaf \in \tau_\x]} \Bigg) \,
\mathrm{d} \Theta \, \mathrm{d}{\bm{w}}.
%\end{equation}
%
\label{eq:BayesScorePCsubst}
\end{align}
Since we assume parameter-consistent determinism, the induced tree $\tau_\x$ remains fixed when ranging over $\Theta,\bm{w}$.
Together with parameter independence, this allows us to pull the integrals into the products, yielding 
\begin{align}
\mathcal{B}_{PC}(\graph)
= & \prod_{\sumnode \in V} \left( \int {1 \over B(\bm{\alpha}_{\sumnode})}  \prod_{\node \in \inputs(\sumnode)}  w_{\sumnode \node}^{\alpha_{\sumnode\node} + n[\sumnode \node] - 1}  \mathrm{d}{\bm{w}_\sumnode} \right)  \nonumber \\
& \prod_{\leaf \in V} \left( \int p(\theta_\leaf) \, \prod_{\x \in \data_\leaf} p(\x \cbar \theta_\leaf) \, \mathrm{d}\theta_\leaf \right),
\label{eq:BayesScore_PC}
\end{align}
where $n[\sumnode\node] \coloneqq \sum_{\x \in \gD} \mathds{1}[\sumnode\node \in \tau_\x]$ 
counts how often the edge between $\sumnode$ and $\node$ appears in an induced tree throughout the dataset, and 
\mbox{$\data_\leaf \coloneqq \{ \x \in \data \cbar \leaf \in \tau_\x \}$}
is the sub-dataset of samples where $\leaf$ appears in the corresponding induced tree.

The first line of (\ref{eq:BayesScore_PC}) takes essentially the same form as eq.~(23) in \shortciteA{heckerman1995learning} and yields the widely know \emph{Bayes-Dirichlet} score 
\begin{equation}
\prod_{\sumnode \in V} \left( {\Gamma(\alpha_{\sumnode}) \over \Gamma(n[\sumnode] + \alpha_{\sumnode})} 
\prod_{\node \in \inputs(\sumnode)} {\Gamma(n[\sumnode \node] + \alpha_{\sumnode\node}) \over \Gamma(\alpha_{\sumnode\node})} \right), 
\end{equation}
where 
$\alpha_{\sumnode} \coloneqq \sum_{\node \in \inputs(\sumnode)} \alpha_{\sumnode\node}$ and 
$n[\sumnode] \coloneqq \sum_{\node \in \inputs(\sumnode)} n[\sumnode \node]$.
The factors in the second line of (\ref{eq:BayesScore_PC}) are exactly eq.~(\ref{eq:leaf_Bayes}), which we assumed to be tractable.
For instance, if the leaves are categorical, the solution is again the Bayes-Dirichlet score; if they are Gaussian, we can use the Bayesian-Gaussian score \shortcite{geiger1994learning}.

\section{BAYESIAN INFORMATION CRITERION}
\label{sec:BIC}

As an alternative to the Bayesian score, we also adapt the \emph{Bayesian information criterion} (BIC) as a structure score for PCs. 
In general, given a graph structure $\graph$ and a training set $\gD$, the BIC score is defined as
\begin{equation}
    \mathrm{BIC}(\graph) = \mathrm{LL}(\graph; \gD) - \frac{\log \lvert \gD \rvert}{2} \lVert \graph \rVert,
\end{equation}
where $\mathrm{LL}(\graph; \data)$ is the log-likelihood under the graph structure $\graph$ when using maximum likelihood parameters (potentially with Laplace smoothing), $\lvert \data \rvert$ is the size of $\data$ and $\lVert \graph \rVert$ is the number of independent parameters encoded in $\graph$. The second term $\frac{\log \lvert \data \rvert}{2} \lVert \graph \rVert$ is essentially a regularization term penalizing complex structures.
BIC is often used as an alternative to BD and for Bayesian networks its asymptotic consistency has been shown \shortcite{koller2009probabilistic}.

The BIC score has been used to learn CNets in \shortciteA{dimauro2015learning}. 
However, the authors defined $\lVert \graph \rVert$ as the number of \emph{conditioning} nodes in the CNet, but did \emph{not} count the parameters of the contained Chow-Liu trees (see next section for details), which actually undercounts the independent parameters in the corresponding PC and weakens the penalty. 
In our experiments, we show that the corrected BIC score reflects better generalization performance and outperforms \shortciteA{dimauro2015learning}.

\section{SCORE-BASED CUTSET LEARNING}
\label{sec:cnetlearning}

Ideally, we would like to use our structure scores to find a \emph{global optimum} among all deterministic PCs, which is presumably NP-hard.
Thus, in order to use our scores for structure learning, we need a method to ``navigate'' through the space of deterministic PCs. 
\emph{Cutset learning} \shortcite{rahman2014cutset} is an attractive method to this end, since it performs greedy search using large structural changes, making it fast and effective.
In this section, we provide the required background and adapt cutset learning to our purposes, i.e., guide the search by our structure scores.
We will for simplicity consider \emph{binary} data, although cutset learning can be generalized to continuous data as well.

\noindent 
\textbf{Chow-Liu Trees.} 
A Chow-Liu tree (CLT) is a Bayesian network \shortcite{koller2009probabilistic}, denoted as $(\gT, \Theta)$ where $\gT$ is a directed tree (meaning each node has at most one parent) over random variables $\X$ and \mbox{$\Theta = \{\theta_{X|\pa(X)}\}_{X \in \X}$} is a collection of conditional probability tables (CPTs), where $\pa(X)$ denotes the parent of $X$ in $\gT$.
The Chow-Liu Algorithm \shortcite{chow1968approximating} learns the tree structure $\gT$ by running a maximum spanning tree algorithm \shortcite{kruskal1956shortest} on the fully-connected graph over $\X$, weighted with the mutual information between pairs of variables, which is estimated from data.
As shown by Chow \& Liu, this spanning tree yields a maximizer of the \emph{log-likelihood structure score} \shortcite{koller2009probabilistic}, making the Chow-Liu algorithm an early example of a principled structure learning algorithm.
The edges of maximal spanning tree are then directed and $\Theta$ is estimated using maximum likelihood.

\noindent 
\textbf{Cutset Networks.} 
Cutset networks (CNets) \shortcite{rahman2014cutset} improve CLTs by embedding them in a hierarchical conditioning process. 
A CNet is a binary decision tree, whose decision nodes correspond to some variable in $\X$ and whose leaves are CLTs over the undecided variables, i.e., the variables not appearing in any decision node on the unique path from root to leaf.
Further, the outgoing edges of decision nodes are equipped with normalized weights.
For any sample $\x$, the probability assigned by the CNet is the product of weights on the path from root to the selected CLT leaf (following decisions according to the values in~$\x$), times the probability the CLT assigns to the undecided variables.
An example CNet is shown in Figure~\ref{fig:cnet}.

\cnetfigure
\algstruclearning

CNets have been extensively studied in literature since they deliver simple, effective and fast structure learning algorithms \shortcite{rahman2014cutset, dimauro2015learning, dimauro2017fast, dang2020strudel, di2021random}.
CNets can be converted into \emph{smooth, deterministic and decomposable PCs} \shortcite{dang2020strudel, di2021random}, making them amenable to our structure scores.

Pseudo code for cutset learning with structure scores is shown in 
Algorithm~\ref{alg:structurelearning}, which recursively selects variables to condition upon.
Our strategy is to select a variable, which, when used in a new decision node with two conditional CLT leaves, would increase our score most.
This selection is done in \textsc{SelectBestCut}, which uses either our Bayesian or BIC score of the corresponding PC.\footnote{In our implementation we avoided explicit compilation to PCs, but performed the corresponding computations directly within the CNet.}
As testing every unconditioned variable increases runtime linearly in the overall number of variables, we restrict the selection to the set of the $\lambda$ most promising candidates (\mbox{\textsc{SelectBestCandidates}}) according to the information gain heuristic in \shortciteA{rahman2014cutset}.
This heuristic, akin to \emph{beam search}, did not impact performance but yielded substantial runtime improvements.
Our structure learner stops as soon as no improvement of the score is achieved.

\section{STRUCTURAL EXPECTATION-MAXIMIZATION}

Since the Bayesian structure score is the (marginal) likelihood of the structure, it can naturally be incorporated in larger probabilistic frameworks such as structural expectation-maximization \shortcite{DBLP:conf/uai/Friedman98}.
Specifically, we consider mixtures of CNets of the form 
$$
p_{mix}(\x) = \sum_{k=1}^K a_k \, p(\x \cbar \Theta, \bm{w}, \graph)
$$
where $a_k \geq 0$ and $\sum_k a_k = 1$, and $K$ is the number of components.
We initialize the model by clustering the data set into $K$ clusters using k-means, and training individual CNets on each portion.
The structural E-step consists of computing the responsibilities of each component proportional to $\gamma_k \propto a_k \, p(\x \cbar \graph)$, where we use the posterior predictive distribution to average over parameters. 
The responsibilities are then used in the structural M-step as (i) weighted average to update $a_k$ and (ii) as fractional samples within our cutset learner.
It should be noted that also other (heuristic) structure learners can be employed in such a scheme, as they presumably also optimize the structure likelihood in some indirect way.

\section{EXPERIMENTS}
We evaluate our approach on both density estimation and image completion tasks, where for simplicity we focus on binary data.
We use the full Bayesian score---yielding the BDeu score \shortcite{heckerman1995learning} in the context of binary data, denoted as CNetBD---and the BIC score, denoted as CNetBIC. 
We limit the number $\lambda$ of candidate conditioning nodes in these two approaches to ten.

The following structure learners of PCs from literature are used as competitors: the original cutset network (CNet) \shortcite{rahman2014cutset}, XCNet \shortcite{dimauro2017fast}, dCSN \shortcite{dimauro2015learning}, Strudel \shortcite{dang2020strudel}, LearnSPN \shortcite{gens2013learning} and ID-SPN \shortcite{rooshenas2014learning}. 
All cutset learners except for dCSN are developed in Python using the
DeeProb-kit library \shortcite{loconte2022deeprob}. 
For LearnSPN, we report results from \shortcite{gens2013learning}. 
For all other structure learners, we use\ the open-source implementations from respective authors.

\begin{figure*}[p]
\begin{center}
\includegraphics[width=\linewidth]{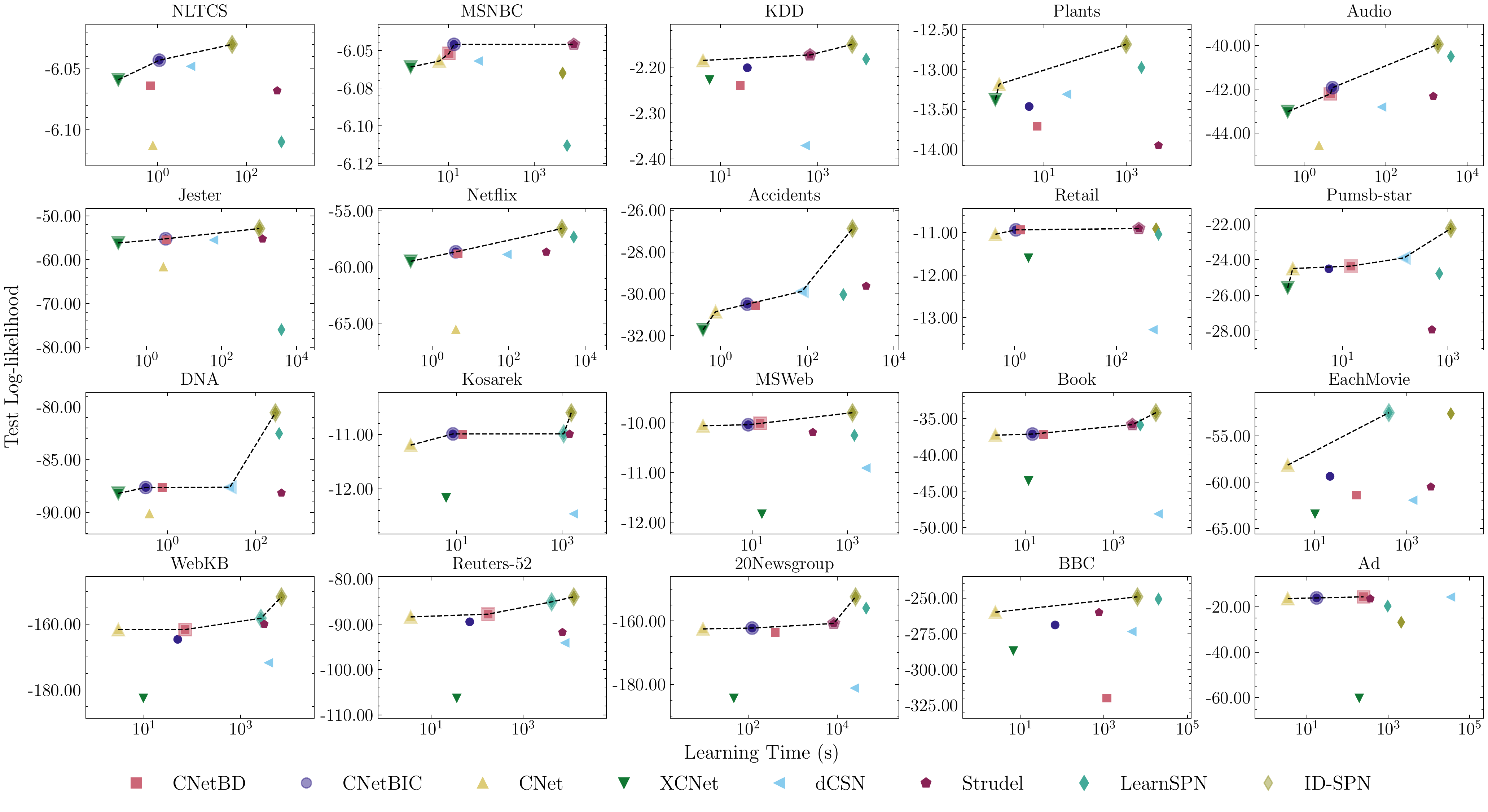}
\caption{Test log-likelihoods (in nats) vs.~learning times of all structure learners on 20 benchmark data sets. 
The optimal learners on the Pareto curve are highlighted with amplified "aura" markers.}
\label{fig:tradeoff}    
\end{center}
\end{figure*}

\subsection{Standard Density Estimation Benchmarks}
As a first experiment, we evaluate all structure learning approaches as density estimators on twenty commonly-used benchmark data sets.
For CNetBD and CNetBIC, the equivalent sample size (ESS, uniform Dirichlet parameter $\alpha$) and the Laplace smoothing factor ($\beta$) are the only hyper-parameters, respectively. 
For all data sets, we fix $\alpha = 0.1$ for CNetBD and $\beta = 0.01$ for CNetBIC, rendering our methods are de facto hyper-parameter-free.

Moreover, the Bayesian nature of our scores should effectively protect against overfitting and directly correspond to generalization.
Hence, to demonstrate this benefit, we train on all available data, without the need of a separate validation set. 
We apply the same procedure (fixed hyperparameter, no validation set) to the more simple structure learners CNet, XCNet and dCSN, and provide the more ``sophisticated'' structure learners Strudel, LearnSPN, and ID-SPN with a validation set.

Specifically, we re-train CNet on all data sets using the hyper-parameters reported in \shortciteA{rahman2014cutset}. 
In the case of XCNet and dCSN, we set $\beta = 0.5$, $\delta = 500$ (the minimum number of samples to decompose) and $\xi = 3$ (the minimum number of features to decompose). 
These hyper-parameter values are expected to perform well on all data sets and are used as default values in their open-source implementations. 
Due to the randomness of XCNet, we use the average test log-likelihood and report the average time on ten different runs.

The other structure learners---Strudel, LearnSPN and ID-SPN---have more complex learning strategies. 
Their implementations or reported results all rely on cross-tuning on a separate validation set. 
Therefore, we use the original training and validation splits for these methods, rather than the combined data.

\tablls

\figuresamples

Ideally, we wish to learn density estimators that are both accurate in terms of the test log-likelihood and efficient in terms of the training time. 
Therefore, we consider the trade-off between the learning time and the accuracy as a multi-objective optimization problem. 
\autoref{fig:tradeoff} shows visualizations of the learning time and test log-likelihoods of all structure learners for each data set. We plot a Pareto frontier as a dashed line for each data set. 
The Pareto-efficient methods on the line are emphasized with amplified ``aura'' markers. From \autoref{fig:tradeoff}, we observe that both CNetBD and CNetBIC appear to be Pareto-efficient on most data sets. 
Compared to other cutset learners with the same greedy learning strategy, CNetBD and CNetBIC achieved leading performance among fast methods. 
In comparison with more complex learners, CNet and CNetBIC still outperform or are on par with Strudel and LearnSPN on many data sets.

The detailed results for each method are summarized in \autoref{tab:ll_models}.
Here we also include the mixture model based learned with structural EM (EM-CNetBD) to compare with the ``complex structure learners.''
Since the mixture model has the number of mixture components as a hyper-parameter, we used the same split in training and validation sets as used for the complex learners, and considered $K \in \{2, 3, 5, 8, 10, 20\}$.
We see that mixtures learned with structural EM consistently improved over single models.
Moreover, our mixtures are close to or improve state-of-the-art PCs.
Additional results can be found in Appendix~\ref{sec:additional_results}.

\subsection{Binary MNIST}
We further evaluate our models on the Binary-MNIST data set \shortcite{larochelle2011neural}. We compare against Einsum networks (EiNets) \shortcite{peharz2020einsum}, which allows large-scale training due to its tensor-based implementation. 
CNetBD (50K parameters) achieves a test log-likelihood of $-118.71$, and a mixture model based on CNetBD with 300 components (470K parameters) has a test log-likelihood of $-103.95$. 
In comparison, an EinNet generally has an rather large number of parameters, e.g.~EinNet (1.2M parameters) has a test log-likelihood of $-113.55$ and a larger EinNet (84M parameters) achieves $-99.82$.

\autoref{fig:samples} shows image generation from CNetBD and EM models with different numbers of components. 
We can clearly observe that the generated images become more plausible with increasing the number of components. 
Moreover, \autoref{fig:samples} shows in inpainting experiments using approximate MPE inference \shortcite{poon2011sum}, where we can see the reconstructions of covered images from mixture models are plausible.

\section{CONCLUSION}

Structure learning is a challenging task in PCs. 
While many techniques have been proposed in recent years, most of them are based on heuristics rather than on principled learning objectives. This is in stark contrast to classical PGMs, where constraint-based and score-based approaches have been used.
Our main motivation in this paper is to establish such principled approaches also for PC structure learning.

In particular, we derived the Bayesian score known from the PGM literature to structure learning of \emph{deterministic} PCs, and furthermore proposed a correction to the BIC score for PCs. 
Both scores were used in a simple structure learner and were allowed to ``speak for themselves:'' no extensive cross-validation and no other defenses against overfitting, such as early stopping, were required.
The results demonstrate that such a simple and principled approach can be effective and efficient.
We hope that our results stimulate further research in this direction.

Our approach is appealing both from a theoretical and practical perspective, since our score is derived from probabilistic considerations, while being analytically tractable and fast in practice.
The main caveat is that our score is restricted to \emph{deterministic} circuits.
Future directions, however, might include approximations of these scores to general (non-deterministic) PC, using e.g.~sampling \shortcite{vergari2019automatic,trapp2019bayesian}, variational techniques \shortcite{zhao2016collapsed}, continuous relaxations \shortcite{pmlr-v181-lang22a}, or numeric integration \shortcite{correia2022continuous}.

Further interesting directions are studying the theoretical properties of Bayesian scores for PCs.
For example, for PGMs, consistency of both the Bayesian and BIC score has been shown, meaning that they will identify the correct structure in the infinite data limit, provided that there exists a PGM which is a perfect map for the data-generating distribution \shortcite{koller2009probabilistic}.
It is likely that such properties also hold for PCs, which might explain the similar performance of the BD and BIC scores in our experiments.
However, rigorously establishing these results is left to future work.

\subsubsection*{Acknowledgements}
We thank the Eindhoven Artificial Intelligence Systems Institute (EAISI) for its support.
This research was supported by the Graz Center for Machine Learning (GraML).
We thank YooJung Choi, Antonio Vergari, and Martin Trapp for helpful discussions.

\bibliographystyle{apacite}
\bibliography{references}

\appendix
\onecolumn

\section{Learning a Chow-Liu Tree}\label{sec:learn_clt}
The algorithm \textsc{LearnCLT} for learning a Chow-Liu tree is shown in Algorithm~\ref{alg:clt}. Overall, for a dataset $\gD$, the algorithm has time complexity $O(|\X|^2 |\gD|)$.

\algclt

\section{Candidate Selection Heuristic}\label{sec:selection_heuristic}
We adapt the information gain heuristic introduced by \shortciteA{rahman2014cutset} to restrict the conditioning node selection to a fixed number of candidates. Intuitively, the heuristic wishes to select variables with maximum information gain or expected entropy reduction by conditioning on the variables. Concretely, the entropy $\hat{H}(\gD)$ of a data set $\gD = \{\x^{(n)}\}_{n=1}^{N}$ over random variables $\rmX$ is defined as
\begin{equation}
    \hat{H}(\gD) = \frac{1}{\lvert \rmX \rvert} \sum_{X \in \rmX} H_\gD(X),
\end{equation}
where $H_\gD(X)$ is the entropy of RV $X$ given $\gD$, which is given by
\begin{equation}
    H_\gD(X) = -\sum_{x \in \gX} p(x)\log p(x),
\end{equation}
where $\gX$ is the set of all possible states of $X$ when assuming $X$ is discrete. Based on the entropy of the data, the information gain $\mathrm{Gain}(X_i)$ after conditioning on an RV $X_i$ is defined as
\begin{equation}
    \mathrm{Gain}(X_i) = \hat{H}(\gD) - \sum_{x_i \in \gX_i} \frac{\lvert \gD_{x_i} \rvert}{\lvert \gD \rvert} \hat{H}(\gD_{x_i}),
\end{equation}
where $\gD_{x_i} = \{\x^{(n)} \in \gD \mid x_i^{(n)} = x_i\}$. In \textsc{SelectBestCandidates}, we select top $\lambda$ RVs with the maximum information gain to form a candidate set $\tilde{\mathbf{X}}$.

\section{Additional Experimental Results}
\label{sec:additional_results}

\subsection{Data Sets}
\label{sec:datasets}

The characteristics of the 20 benchmark density estimation data sets are reported in \autoref{tab:info_datasets}. The number of variables ranges from $16$ to $1556$.

\tabdatasetcharacteristics

\subsection{Learning Time}
\label{sec:learning_time}

To evaluate the efficiency of different structure learners, we report their learning times in \autoref{tab:time_models}, which are all measured in total seconds during training. Combining with the test log-likelihoods of various structure learners, we can observe both CNetBD and CNetBIC achieve a better trade-off between the performance and the training time.

\tablearningtimes

\subsection{Circuit Size}\label{sec:circuit_size}
\autoref{tab:circuit_sizes} summarizes the circuit size of models learned by different cutset structure learners.
Clearly, CNetBD and CNetBIC deliver much smaller circuits than other cutset learners on almost all data sets.
In addition, compare to other cutset learners, CNetBD and CNetBIC often achieve leading performance.
These further validate our theoretical analysis that both the BDeu and the BIC score can effectively protect against overfitting. 

\tabcircuitsize

\vfill

\end{document}

%% file: tables.tex
\newcommand{\tabdatasetcharacteristics}{
\begin{table}
  \centering
  \scriptsize
  \begin{tabular}{lrrrr}
  \toprule
Dataset & \#Features & \#Train & \#Valid. & \#Test \\ \hline
NLTCS & 16 & 16181 & 2157 & 3236 \\
MSNBC & 17 & 291326 & 38843 & 58265 \\
KDD & 64 & 180092 & 19907 & 34955 \\
Plants & 69 & 17412 & 2321 & 3482 \\
Audio & 100 & 15000 & 2000 & 3000 \\
Jester & 100 & 9000 & 1000 & 4116 \\
Netflix & 100 & 15000 & 2000 & 3000 \\
Accidents & 111 & 12758 & 1700 & 2551 \\
Retail & 135 & 22041 & 2938 & 4408 \\
Pumsb-star & 163 & 12262 & 1635 & 2452 \\
DNA & 180 & 1600 & 400 & 1186 \\
Kosarek & 190 & 33375 & 4450 & 6675 \\
MSWeb & 294 & 29441 & 3270 & 5000 \\
Book & 500 & 8700 & 1159 & 1739 \\
EachMovie & 500 & 4525 & 1002 & 591 \\
WebKB & 839 & 2803 & 558 & 838 \\
Reuters-52 & 889 & 6532 & 1028 & 1540 \\
20Newsgroup & 910 & 11293 & 3764 & 3764 \\
BBC & 1058 & 1670 & 225 & 330 \\
Ad & 1556 & 2461 & 327 & 491 \\
  \bottomrule
  \end{tabular}
  \caption{Characteristics of the benchmark density estimation data sets}
  \label{tab:info_datasets}
\end{table}
}

\newcommand{\tablls}{
\begin{table*}[p]
  \centering
  \tiny
  \begin{tabular}{lrrrrr|rrrr}
  \toprule
Data Set    & CNet                 & XCNet    & dCSN                & CNetBD              & CNetBIC              & Strudel             & LearnSPN             & ID-SPN              & EM-CNetBD            \\ \hline
NLTCS       & -6.113               & -6.059   & {-6.048}  & -6.064              & \textbf{-6.043}      & -6.068              & -6.110               & \textbf{-6.030}     & {-6.032}   \\
MSNBC       & -6.057               & -6.061   & -6.057              & {-6.052}  & \textbf{-6.046}      & {-6.046}  & -6.113               & -6.065              & \textbf{-6.041}      \\
KDD         & \textbf{-2.185}      & -2.227   & -2.371              & -2.240              & {-2.201}   & -2.173              & -2.182               & \textbf{-2.150}     & {-2.159}   \\
Plants      & \textbf{-13.187}     & -13.381  & {-13.312} & -13.713             & -13.466              & -13.956             & -12.977              & \textbf{-12.687}    & {-12.902}  \\
Audio       & -44.571              & -43.022  & -42.810             & {-42.210} & \textbf{-41.927}     & -42.319             & -40.503              & {-39.950} & \textbf{-39.943}     \\
Jester      & -61.670              & -56.173  & -55.506             & {-55.454} & \textbf{-55.231}     & -55.255             & -75.989              & \textbf{-52.886}    & \textbf{-52.923}     \\
Netflix     & -65.572              & -59.492  & -58.877             & {-58.820} & \textbf{-58.649}     & -58.659             & -57.328              & \textbf{-56.575}    & {-56.604}  \\
Accidents   & -30.860              & -31.720  & \textbf{-29.880}    & -30.568             & {-30.497}  & {-29.633} & -30.038              & \textbf{-26.876}    & -29.919              \\
Retail      & -11.043              & -11.595  & -13.282             & {-10.940} & \textbf{-10.939}     & -10.908             & -11.043              & \textbf{-10.913}    & {-10.927}  \\
Pumsb-star  & -24.497              & -25.575  & \textbf{-23.900}    & {-24.361} & -24.513              & -27.935             & -24.781              & \textbf{-22.251}    & {-24.009}  \\
DNA         & -90.143              & -88.199  & \textbf{-87.621}    & -87.643             & {-87.642}  & -88.167             & {-82.523}  & \textbf{-80.550}    & -85.437              \\
Kosarek     & -11.197              & -12.165  & -12.462             & {-10.999} & \textbf{-10.991}     & -10.993             & -10.989              & \textbf{-10.600}    & {-10.695}  \\
MSWeb       & -10.060              & -11.830  & -10.909             & \textbf{-10.012}    & {-10.039}  & -10.192             & -10.252              & \textbf{-9.797}     & {-9.963}   \\
Book        & -37.293              & -43.550  & -48.116             & {-37.181} & \textbf{-37.143}     & -35.841             & -35.886              & \textbf{-34.200}    & {-34.854}  \\
EachMovie   & \textbf{-58.153}     & -63.420  & -61.946             & -61.400             & {-59.365}  & -60.503             & \textbf{-52.485}     & {-52.588} & -53.378              \\
WebKB       & {-161.650} & -182.453 & -171.743            & \textbf{-161.634}   & -164.611             & -159.989            & -158.204             & \textbf{-151.680}   & {-157.053} \\
Reuters-52  & {-88.386}  & -106.251 & -94.102             & \textbf{-87.766}    & -89.461              & -91.778             & -85.067              & \textbf{-83.954}    & {-84.993}  \\
20Newsgroup & {-162.488} & -184.297 & -181.165            & -163.671            & \textbf{-162.227}    & -160.786            & -155.925             & \textbf{-152.37}    & {-155.002} \\
BBC         & \textbf{-259.870}    & -286.869 & -273.465            & -320.052            & {-268.861} & -260.084            & {-250.687} & \textbf{-249.12}    & -258.691             \\
Ad          & -16.436              & -60.120  & {-15.774} & \textbf{-15.676}    & -16.161              & {-16.521} & -19.733              & -26.85              & \textbf{-14.923}     \\
\bottomrule
  \end{tabular}
  \caption{Average test log-likelihoods (in nats) for all structure learners. The cutset learners are in the left group and other structure learners are in the right group. The model with the best performance in each group is highlighted in bold.}
  \label{tab:ll_models}
\end{table*}
}

\newcommand{\tablearningtimes}{
\begin{table*}
  \centering
  \tiny
  \begin{tabular}{lrrrrr|rrr}
  \toprule
Data Set    & CNet & XCNet  & dCSN    & CNetBD & CNetBIC & Strudel & LearnSPN & ID-SPN \\ \hline
NLTCS       & 0.8 & 0.1   & 5.6     & 0.7    & 1.1   & 500  & 623   & 48    \\
MSNBC       & 6.1 & 1.3   & 50.2    & 10.3   & 13.5  & 8374 & 5843  & 4604  \\
KDD         & 4.4 & 6.0   & 563.4   & 25.6   & 35.7  & 697  & 9943  & 5126  \\
Plants      & 0.8 & 0.7   & 35.9    & 6.8    & 4.4   & 5999 & 2318  & 986   \\
Audio       & 2.3 & 0.4   & 80.8    & 4.4    & 4.9   & 1476 & 3977  & 1920  \\
Jester      & 2.8 & 0.2   & 62.6    & 3.3    & 3.3   & 1239 & 3946  & 1020  \\
Netflix     & 4.2 & 0.3   & 92.0    & 4.7    & 4.1   & 966  & 5062  & 2485  \\
Accidents   & 0.8 & 0.4   & 76.5    & 6.5    & 4.2   & 2284 & 681   & 1102  \\
Retail      & 0.4 & 1.9   & 530.6   & 1.3    & 1.1   & 276  & 671   & 601   \\
Pumsb-star  & 1.1 & 0.9   & 146.4   & 14.5   & 5.5   & 496  & 684   & 1132  \\
DNA         & 0.4 & 0.1   & 26.1    & 0.8    & 0.3   & 382  & 339   & 279   \\
Kosarek     & 1.3 & 6.3   & 1658.7  & 13.1   & 8.5   & 1381 & 1068  & 1485  \\
MSWeb       & 0.9 & 16.1  & 2474.1  & 14.8   & 8.3   & 188  & 1404  & 1279  \\
Book        & 2.1 & 12.0  & 10729.9 & 26.2   & 14.7  & 2742 & 4137  & 9355  \\
EachMovie   & 2.6 & 10.1  & 1360.6  & 79.4   & 21.4  & 3318 & 406   & 9014  \\
WebKB       & 3.0 & 9.8   & 3800.4  & 71.2   & 50.0  & 3069 & 2616  & 6943  \\
Reuters-52  & 3.6 & 36.0  & 8156.3  & 172.6  & 68.9  & 7166 & 4166  & 12653 \\
20Newsgroup & 9.7 & 47.5  & 24660.7 & 403.3  & 122.1 & 8303 & 44341 & 25637 \\
BBC         & 2.5 & 6.8   & 4718.8  & 1175.7 & 68.1  & 765  & 20199 & 6348  \\
Ad          & 3.4 & 193.7 & 34576.8 & 246.5  & 17.3  & 361  & 969   & 2080  \\
\bottomrule
  \end{tabular}
  \caption{Learning times (in seconds) of all structure learners. Due to the randomness of XCNet, we use the average learning time on 10 different runs.}
  \label{tab:time_models}
\end{table*}
}

\newcommand{\tabcircuitsize}{
\begin{table*}
  \centering
  \scriptsize
  \begin{tabular}{lrrrrr}
  \toprule
Data Set    & CNet   & XCNet     & dCSN   & CNetBD & CNetBIC \\ \hline
NLTCS       & 4543   & 1089.4    & 1017   & 205    & 315     \\
MSNBC       & 20143  & 5414.0    & 7069   & 1695   & 1865    \\
KDD         & 8601   & 47638.4   & 47888  & 2823   & 1651    \\
Plants      & 18605  & 21263.8   & 8532   & 3399   & 3163    \\
Audio       & 54651  & 15166.0   & 12530  & 1549   & 3437    \\
Jester      & 72873  & 7037.6    & 6000   & 1359   & 2689    \\
Netflix     & 100497 & 10412.6   & 8258   & 1741   & 2881    \\
Accidents   & 13763  & 15273.6   & 9345   & 2355   & 2775    \\
Retail      & 3667   & 52036.2   & 85730  & 269    & 269     \\
Pumsb-star  & 16023  & 32990.8   & 12203  & 4129   & 3187    \\
DNA         & 5625   & 2329.0    & 1767   & 359    & 359     \\
Kosarek     & 10725  & 175426.0  & 166168 & 1875   & 1875    \\
MSWeb       & 5801   & 455462.6  & 158034 & 2333   & 1753    \\
Book        & 13887  & 219122.2  & 166226 & 2989   & 3981    \\
EachMovie   & 16857  & 185578.2  & 18787  & 8931   & 6959    \\
WebKB       & 10035  & 86246.6   & 16692  & 5023   & 10035   \\
Reuters-52  & 10635  & 334000.0  & 35350  & 10633  & 12403   \\
20Newsgroup & 30783  & 402505.8  & 63209  & 23531  & 19939   \\
BBC         & 6337   & 43752.4   & 12646  & 58849  & 10551   \\
Ad          & 6219   & 1076052.0 & 77207  & 9325   & 3111    \\
\bottomrule
  \end{tabular}
  \caption{Circuit sizes (measured in the number of independent parameters) of all structure learners. Due to the randomness of XCNet, we use the average circuit size on 10 different runs.}
  \label{tab:circuit_sizes}
\end{table*}
}

%% file: figures.tex
\newcommand{\cnetfigure}{
\begin{figure}
\centering
\includegraphics[width=\linewidth]{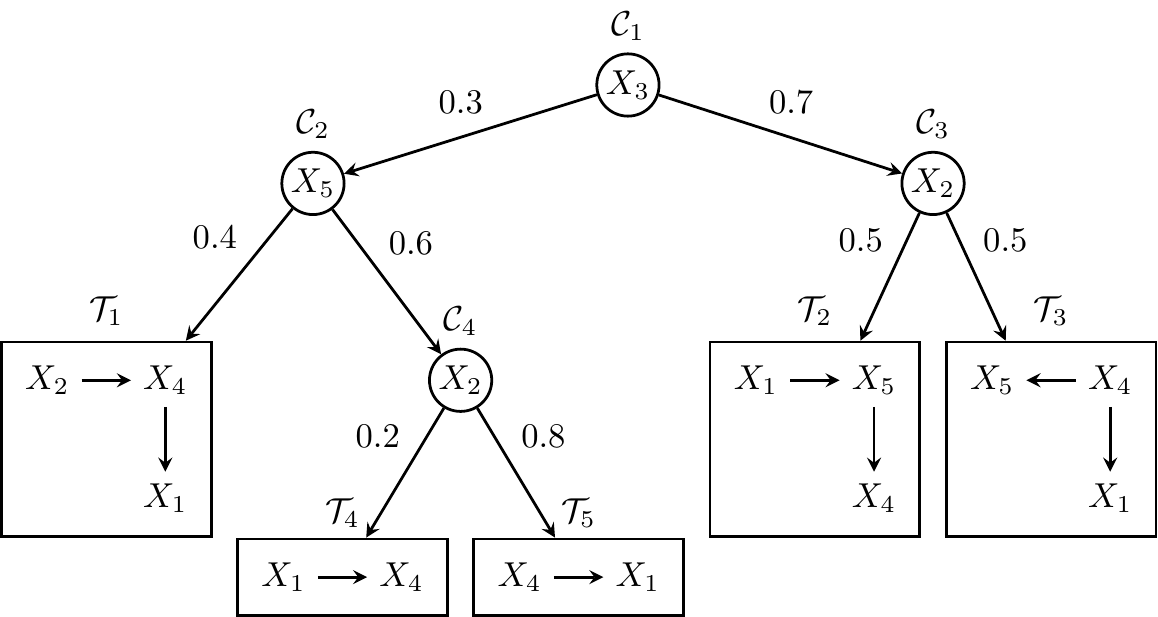}
\caption{A CNet over binary random variables \mbox{$\rmX=\{X_1,X_2,X_3,X_4,X_5\}$}. 
Inner nodes $\{\gC_1,\gC_2,\gC_3,\gC_4\}$ are decision nodes whose left (resp. right) branches represent conditioning on state $0$ (resp. $1$).
The leaves of the CNet are CLTs $\{(\gT_i, \bigtheta_i)\}_{i=1}^{5}$.}
\label{fig:cnet}
\end{figure}
}

\newcommand{\figuresamples}{
\begin{figure*}[p]
\centering
\begin{subfigure}[b]{0.45\linewidth}
\begin{subfigure}[b]{0.32\linewidth}
    \centering
    \includegraphics[width=\linewidth]{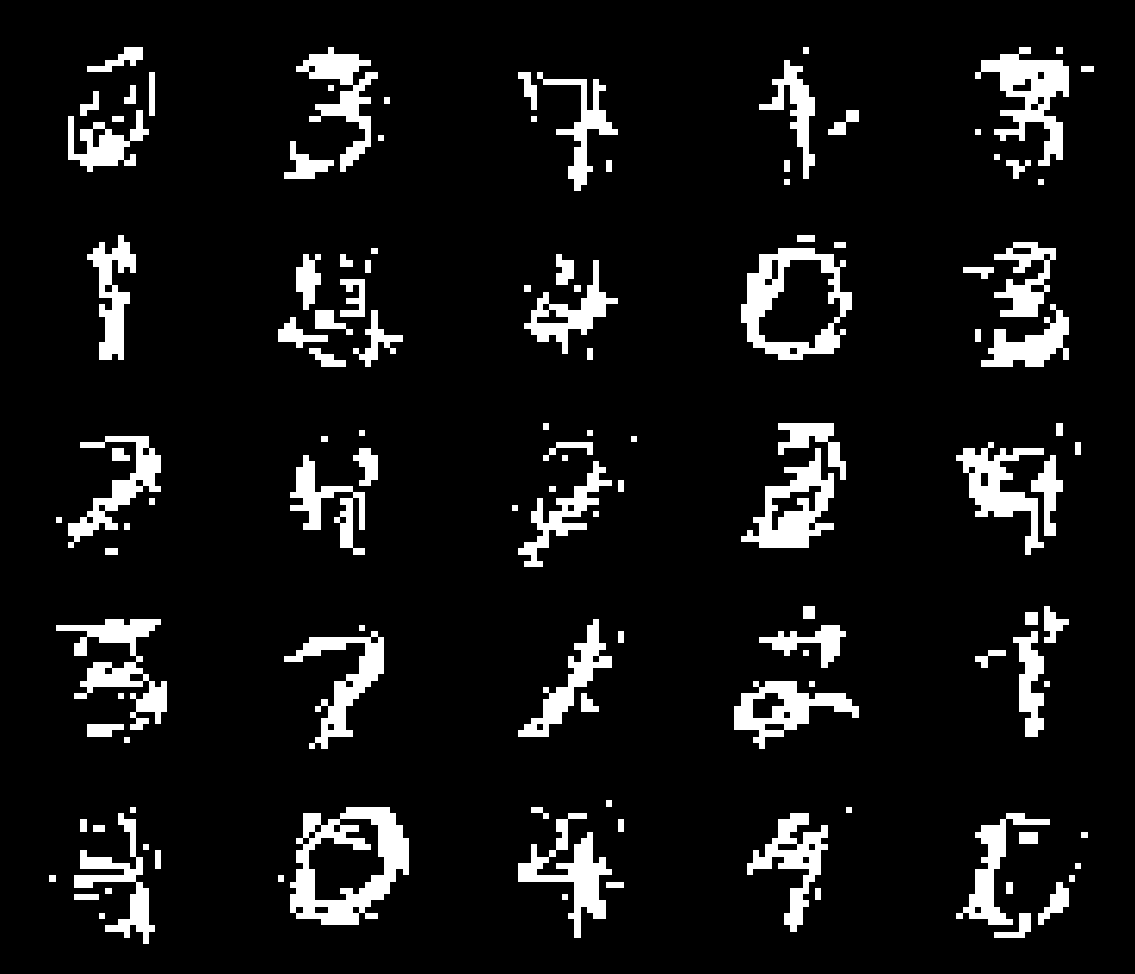}
\end{subfigure}
\begin{subfigure}[b]{0.32\linewidth}
    \centering
    \includegraphics[width=\linewidth]{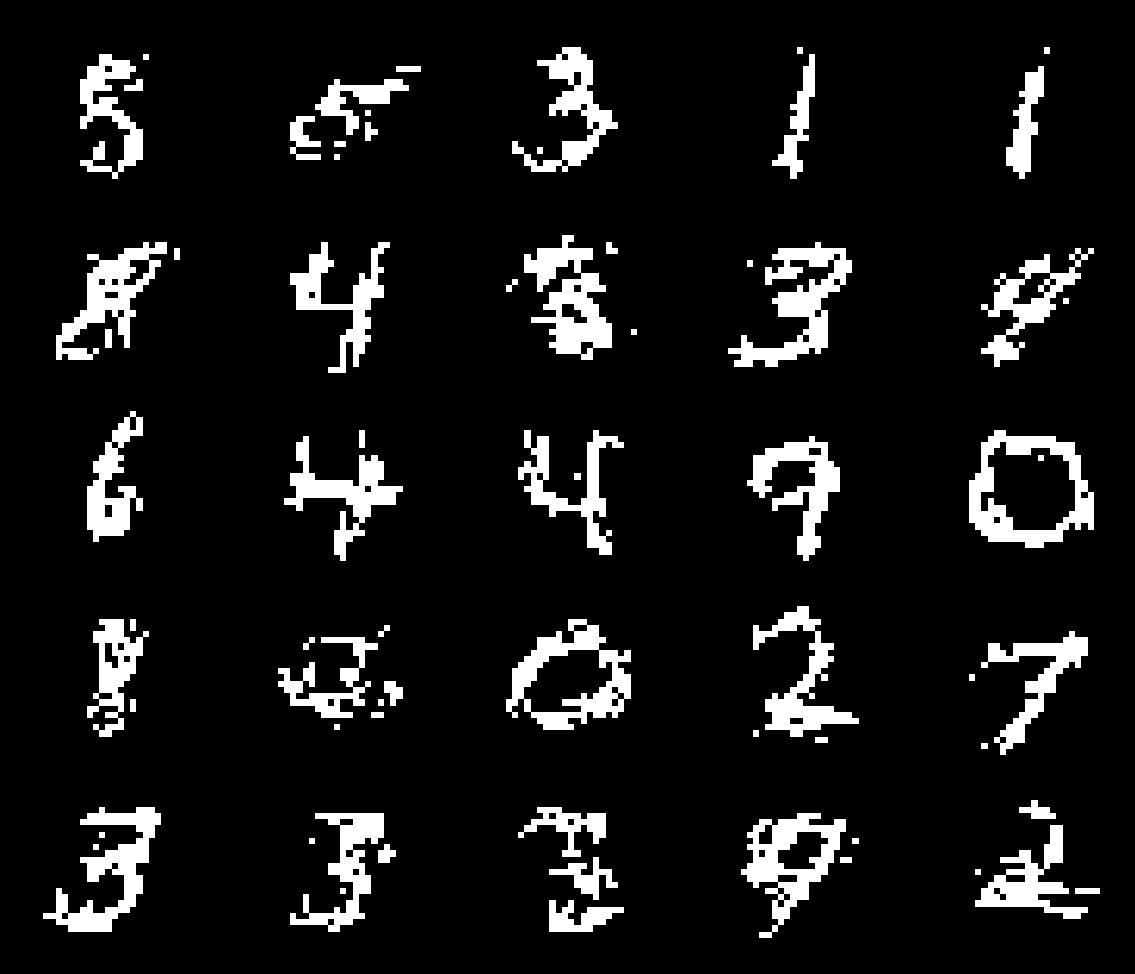}
\end{subfigure}
\begin{subfigure}[b]{0.32\linewidth}
    \centering
    \includegraphics[width=\linewidth]{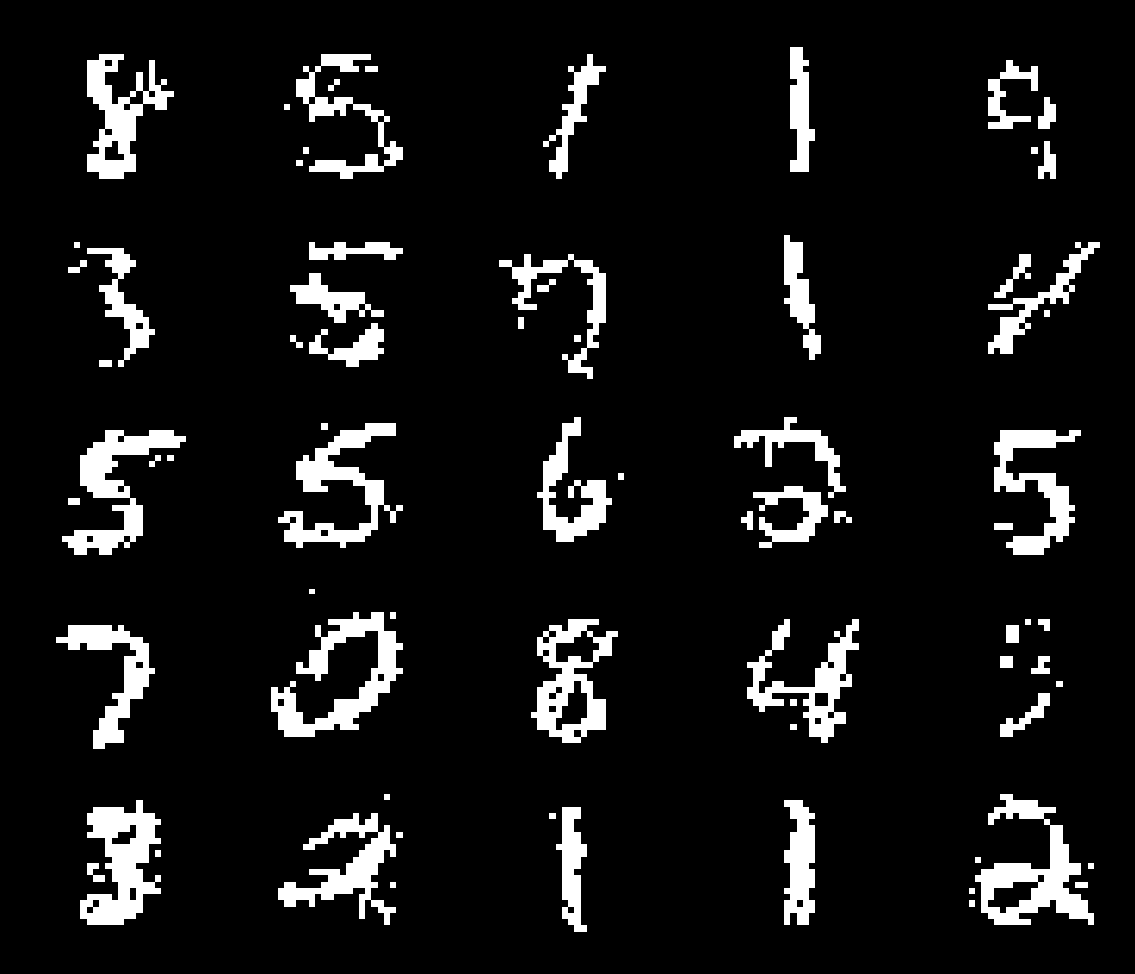}
\end{subfigure}
\caption{Sampling results.}
\end{subfigure}
\begin{subfigure}[b]{0.45\linewidth}
\begin{subfigure}[b]{0.32\linewidth}
    \centering
    \includegraphics[width=\linewidth]{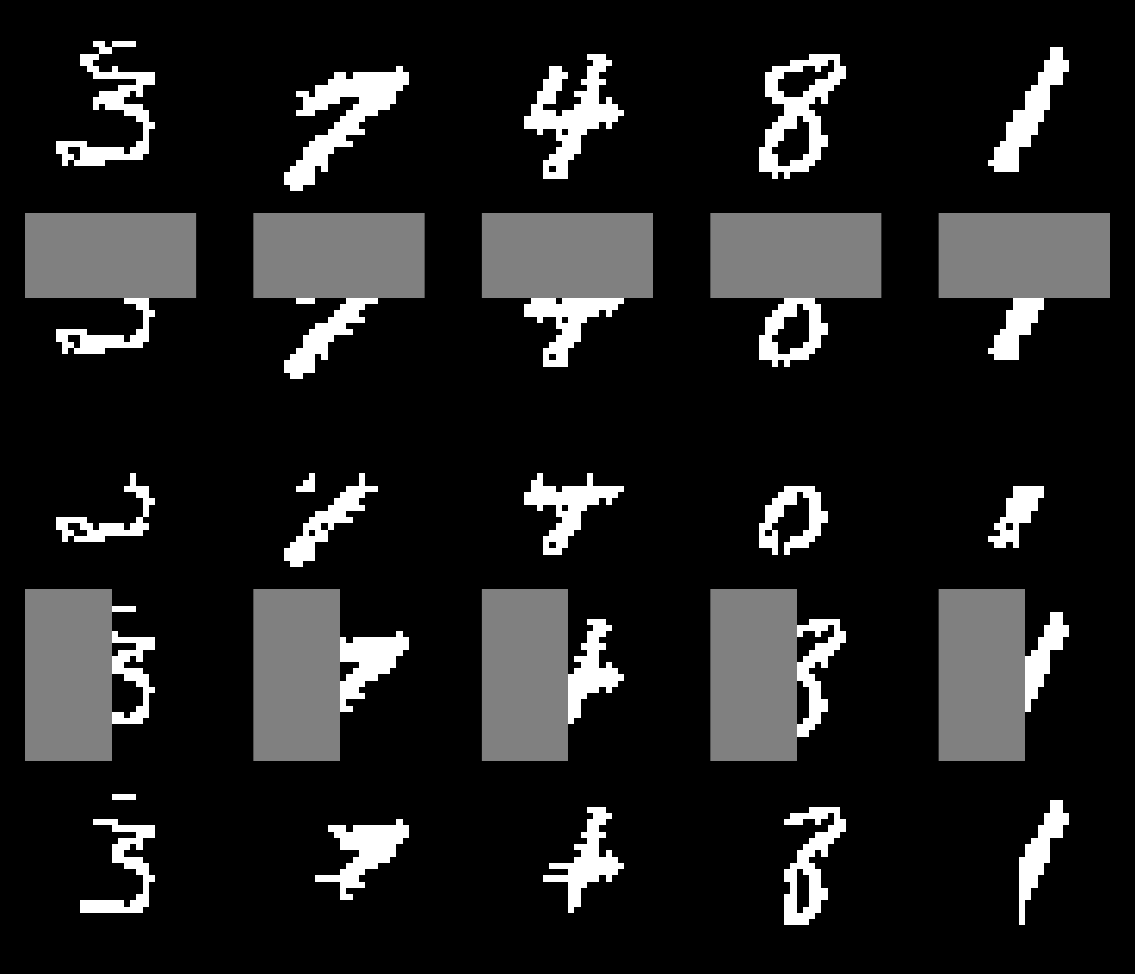}
\end{subfigure}
\begin{subfigure}[b]{0.32\linewidth}
    \centering
    \includegraphics[width=\linewidth]{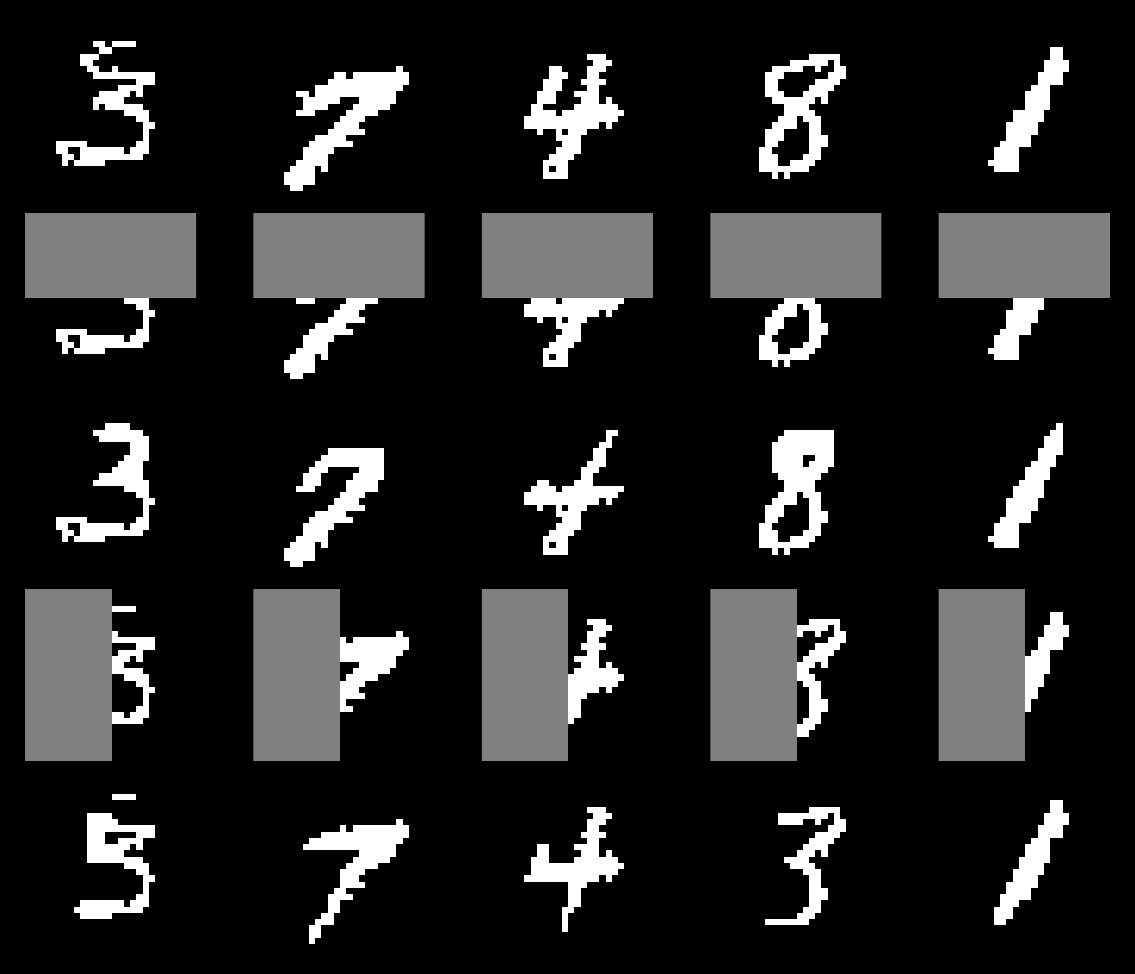}
\end{subfigure}
\begin{subfigure}[b]{0.32\linewidth}
    \centering
    \includegraphics[width=\linewidth]{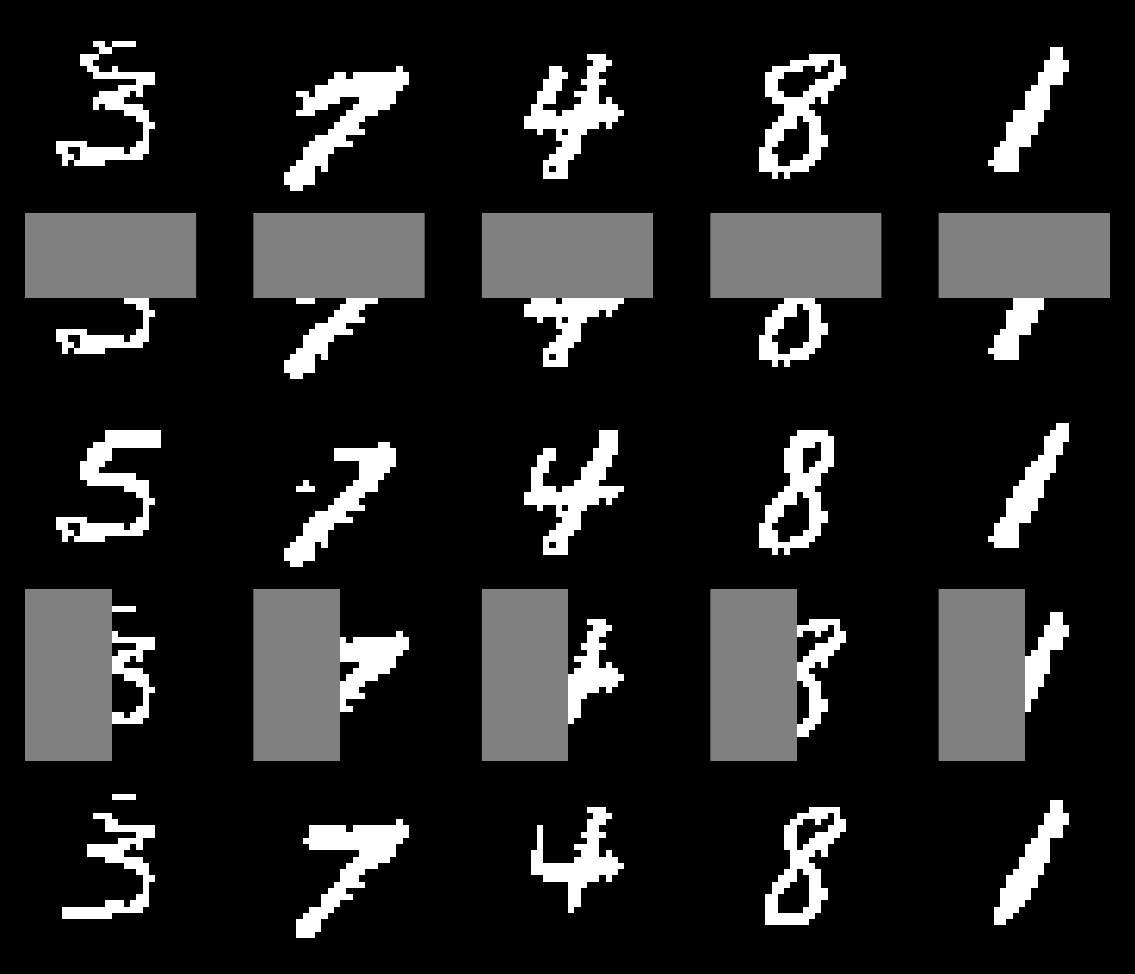}
\end{subfigure}
\caption{Inpainting results.}
\end{subfigure}
\caption{Samples (a) and inpainting results (b) for Binary-MNIST for CNetBD (left), EM-CNetBD with 100 components (middle) and 300 components (right). 
In (b), the original test images are shown in the top row; then in alternating rows the images with missing part and reconstructions are shown.}
\label{fig:samples}
\end{figure*}
}

%% file: algorithms.tex
\newcommand{\algstruclearning}{
\begin{algorithm}[t]
\footnotesize
\caption{\textsc{Cutset Structure Learning}}
\label{alg:structurelearning}
\begin{algorithmic}[1]
\Require A training set $\gD = \{\rvx^{(n)}\}_{n=1}^{N}$ over binary RVs $\mathbf{X}$; number of candidate conditioning nodes $\lambda$.
\Ensure a CNet $\mathcal{C}^*$ representing a deterministic PC.
\vspace*{0.2cm}
\Function{\textsc{Cut}}{$\mathbf{X}, \mathcal{D}, \lambda$}
\State $\mathcal{T}^* \leftarrow \textsc{LearnCLT}(\mathbf{X}, \mathcal{D})$
\State $\tilde{\mathbf{X}} \leftarrow \textsc{SelectBestCandidates}(\mathbf{X}, \mathcal{D}, \lambda)$
\State $X^*, \mathcal{C}^*, \mathcal{D}_l, \mathcal{D}_r \leftarrow \textsc{SelectBestCut}(\tilde{\mathbf{X}}, \mathcal{D})$
\If{$S(\mathcal{T}^*) > S(\mathcal{C}^*$)}
\State \Return $\mathcal{T}^*$
\Else
\State $\mathcal{C}^* \leftarrow \textsc{CNet}(X^*, \textsc{Cut}(\mathbf{X}\backslash X^*, \mathcal{D}_l, \lambda), \newline \hspace*{10.5em} \textsc{Cut}(\mathbf{X}\backslash X^*, \mathcal{D}_r, \lambda))$
\State \Return $\mathcal{C}^*$
\EndIf
\EndFunction
\vspace*{0.2cm}
\Function{\textsc{SelectBestCut}}{$\tilde{\mathbf{X}}, \mathcal{D}$}
\ForAll{$X \in \tilde{\mathbf{X}}$}
\State $\mathcal{D}_l, \mathcal{D}_r \leftarrow \textsc{Split}(X, \mathcal{D})$
\State $\mathcal{T}_l \leftarrow \textsc{LearnCLT}(\mathbf{X}\backslash X, \mathcal{D}_l)$
\State $\mathcal{T}_r \leftarrow \textsc{LearnCLT}(\mathbf{X}\backslash X, \mathcal{D}_r)$
\State $\mathcal{C} \leftarrow \textsc{CNet}(X, \mathcal{T}_l, \mathcal{T}_r)$
\State $S(\mathcal{C}) \leftarrow \textsc{score}(\mathcal{C})$
\EndFor
\State $X^* \leftarrow$ best $X$ with the highest $S(\mathcal{C})$
\State \Return $X^*, \mathcal{C}^*, \mathcal{D}_l, \mathcal{D}_r$
\EndFunction
\end{algorithmic}
\end{algorithm}
}

\newcommand{\algclt}{
\begin{algorithm}
\footnotesize
\caption{\textsc{LearnCLT}}
\label{alg:clt}
\begin{algorithmic}[1]
\Require A training set $\gD = \{\rvx^{(n)}\}_{n=1}^{N}$ over RVs $\mathbf{X}$, equivalent sample size (ESS) $\alpha$.
\Ensure a parameterized Chow-Liu tree $\gT$
\vspace*{0.2cm}
\Function{LearnCLT}{$\gD, \alpha$}
\State $\mathbf{MI} \leftarrow \textsc{ComputeMutualInformation}(\mathbf{X})$
\State $\gT' \leftarrow \textsc{ComputeMaximumSpanningTree}(\mathbf{MI})$
\State $\gT \leftarrow \textsc{AssignEdgeDirections}(\gT')$
\State $\gT \leftarrow \textsc{ComputeBayesianPosteriorParameters}(\gD, \gT, \alpha)$
\State \Return $\gT$
\EndFunction
\end{algorithmic}
\end{algorithm}
}

%% file: math_commands.tex
\usepackage{amsmath,amsfonts,bm}

\newcommand{\X}{\mathbf{X}}
\newcommand{\x}{\mathbf{x}}

\newcommand{\cbar}{\,|\,}
\newcommand{\data}{\mathcal{D}}

\newcommand{\graph}{\mathcal{G}}

\newcommand{\scope}{\sigma}

\newcommand{\inputs}{\mathbf{in}}

\newcommand{\node}{\mathsf{N}}
\newcommand{\sumnode}{\mathsf{S}}
\newcommand{\prodnode}{\mathsf{P}}
\newcommand{\leaf}{\mathsf{L}}

\newcommand{\bigtheta}{\mathbf{\Theta}}
\newcommand{\pa}{\pi}

\def\eqref#1{equation~\ref{#1}}
\def\1{\bm{1}}

\def\rvx{{\mathbf{x}}}

\def\rmX{{\mathbf{X}}}

\DeclareMathAlphabet{\mathsfit}{\encodingdefault}{\sfdefault}{m}{sl}
\SetMathAlphabet{\mathsfit}{bold}{\encodingdefault}{\sfdefault}{bx}{n}

\def\gC{{\mathcal{C}}}
\def\gD{{\mathcal{D}}}

\def\gT{{\mathcal{T}}}

\def\gX{{\mathcal{X}}}